\pdfoutput=1

\documentclass[11pt]{article}

\usepackage{acl}

\usepackage{times}
\usepackage{latexsym}

\usepackage[T1]{fontenc}

\usepackage[utf8]{inputenc}

\usepackage[noend]{algpseudocode}
\usepackage{algorithmicx,algorithm}
\usepackage{bm}
\usepackage{bbm}
\usepackage{microtype}
\usepackage{booktabs}       
\usepackage{amsmath}
\usepackage{amsfonts}
\usepackage{array}
\usepackage{bm}
\usepackage{multirow}
\usepackage{subcaption}
\usepackage{hyperref}       
\usepackage{url}            
\usepackage{graphicx} 
\usepackage{utfsym}
\usepackage{xcolor}
\usepackage{inconsolata}
\usepackage{CJKutf8}
\usepackage{pythonhighlight} 
\usepackage{makecell}
\usepackage[normalem]{ulem}
\title{Are LLMs Aware that\\
Some Questions are not Open-ended?}

\author{Dongjie Yang, 
    Hai Zhao\thanks{\;\;Corresponding author; This paper was partially supported by Joint Research Project of Yangtze River Delta Science and Technology Innovation Community (No. 2022CSJGG1400).
    }\\
Department of Computer Science and Engineering, Shanghai Jiao Tong University, \\Key Laboratory of Shanghai Education Commission for Intelligent Interaction and \\ Cognitive Engineering, Shanghai Jiao Tong University, \\Shanghai Key Laboratory of Trusted Data Circulation and Governance in Web3
    \\
    \texttt{djyang.tony@sjtu.edu.cn,zhaohai@cs.sjtu.edu.cn}
  }

\begin{document}
\maketitle
\begin{abstract}
Large Language Models (LLMs) have shown the impressive capability of answering questions in a wide range of scenarios. However, when LLMs face different types of questions, it is worth exploring whether LLMs are aware that some questions have limited answers and need to respond more deterministically but some do not. We refer to this as \emph{question awareness} of LLMs. The lack of question awareness in LLMs leads to two phenomena that LLMs are: (1) too casual to answer non-open-ended questions or (2) too boring to answer open-ended questions. In this paper, we first evaluate the question awareness in LLMs. The experimental results show that LLMs have the issues of lacking awareness of questions in certain domains, e.g. factual knowledge, resulting in hallucinations during the generation. To mitigate these, we propose a method called Question Awareness Temperature Sampling (QuATS). This method enhances the question awareness of LLMs by adaptively adjusting the output distributions based on question features. The automatic adjustment in QuATS eliminates the need for manual temperature tuning in text generation and consistently improves model performance in various benchmarks.
\end{abstract}

%

\section{Introduction}

Large language models (LLMs) \citep{openai2023chatgpt, openai2023gpt4, anthropic2023claude, jiang2023mistral, bai2023qwen, team2023gemini} have emerged as groundbreaking innovations in achieving a remarkable level of fluency and comprehension in question-answering using the human language \citep{alpaca, vicuna2023, xu2023wizardlm}. Though LLMs can answer enormous questions with their knowledge bases, it is hard to tell if LLMs are aware of the difference between the questions they are answering. In other words, do LLMs understand that, open-ended questions encourage more casual and creative answers, but non-open-ended questions, e.g. problems about calculations and factual knowledge, need more deterministic answers? We refer to this as \emph{question awareness} of LLMs that one knows which type of questions requires deterministic answers and which does not. It is significant to explore the question awareness of LLMs because it has a deep relationship to the model hallucinations that LLMs are prone to generate inaccurate content when they are not sure.

\begin{figure*}[ht]
		\centering
		\includegraphics[width=0.8\linewidth]{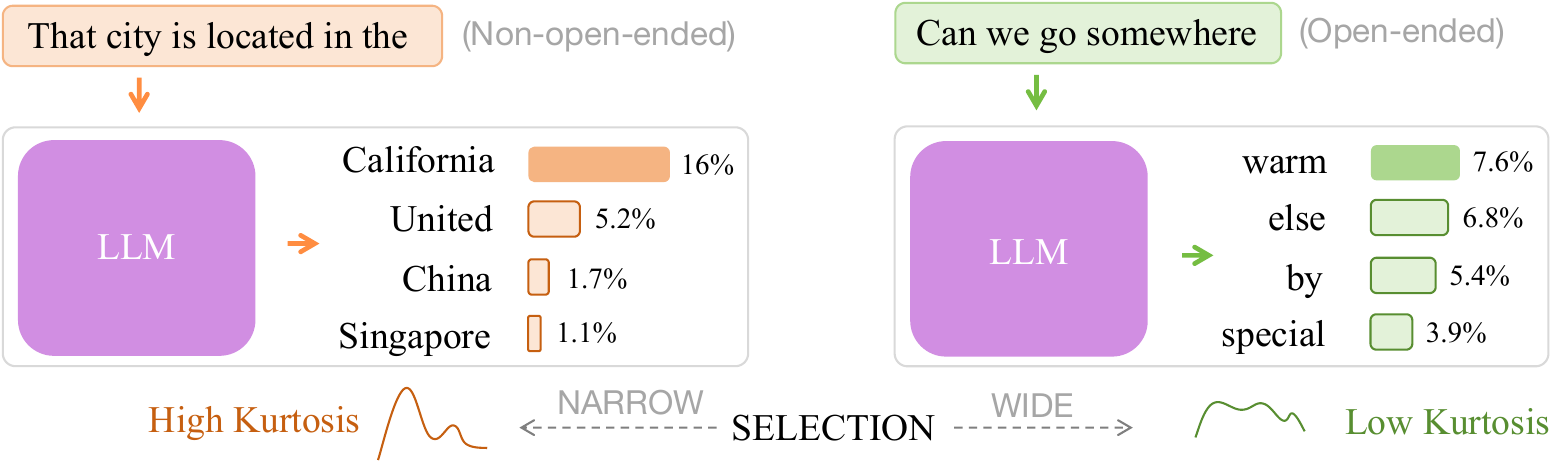}
        \caption{LLMs should choose to be deterministic to answer the question on the left but can have more choices to answer the one on the right.}
		\label{fig:cover}
\end{figure*}

In this paper, we explore whether LLMs have question awareness on different types of questions. Because LLMs sample next tokens from output distributions, as shown in Figure \ref{fig:cover}, we examine the degree of the determinacy of LLMs from the steepness of the output distributions. A steeper output distribution means the model has confidence in selecting which token in the vocabulary to be the next token and a flat one means the model does not have a clear preference for the next token. Therefore, the steepness of the output distributions reflects the question awareness by indicating determinacy about the generated answers. We utilize the kurtosis to measure the steepness of the distribution and investigate the question awareness by checking kurtosises of output distributions when LLMs are asked different types of questions. We evaluate LLaMA 2 \cite{touvron2023llama} and Falcon \citep{penedo2023refinedweb} on different types of non-open-ended/open-ended questions for question awareness evaluation. Experimental results show that LLMs have a certain degree of question awareness but lack the awareness in some scenarios, e.g., factual knowledge, thus easily giving more casual and hallucinated answers. 

As the steepness of output distributions reflects the question awareness, we utilize the temperature of the Softmax function \citep{bridle1989training} to adjust the steepness to externally change the question awareness. We evaluate the model with different temperatures to explore the influence of question awareness on model performance. The results show a relatively lower temperature (steeper distribution) makes the model more deterministic and have better performance on non-open-ended questions.

Inspired by the adjustment of temperature on the question awareness, we propose Question Awareness Temperature Sampling (QuATS), a method that enhances question awareness of LLMs by adjusting the output distributions through the temperature. When facing different questions, LLMs choose to be more deterministic or not using an adaptive temperature strategy of QuATS, avoiding the tedious process of temperature tuning in the text generation. To sum up, our contributions are stated as follows:
\begin{itemize}
    \item We evaluate the question awareness in LLMs and observe that LLMs have the fundamental ability to identify open-ended and non-open-ended questions but lack effective awareness in some domains, e.g., factual knowledge.
    \item We propose Question Awareness Temperature Sampling (QuATS). It enables LLMs to choose to be deterministic or not when answering different questions by adaptively adjusting the temperature without manual tuning.
    \item Our experimental results show that the QuATS enhances the question awareness of LLMs and consistently improves the model performance on various benchmarks.
\end{itemize}

\section{Question Awareness Evaluation}
\label{sec:QuAT_eval}
In this section, we evaluate the question awareness of LLMs and how it influences the model performance on downstream tasks.

\begin{figure*}[ht]
	\centering
	\includegraphics[width=0.9\textwidth]{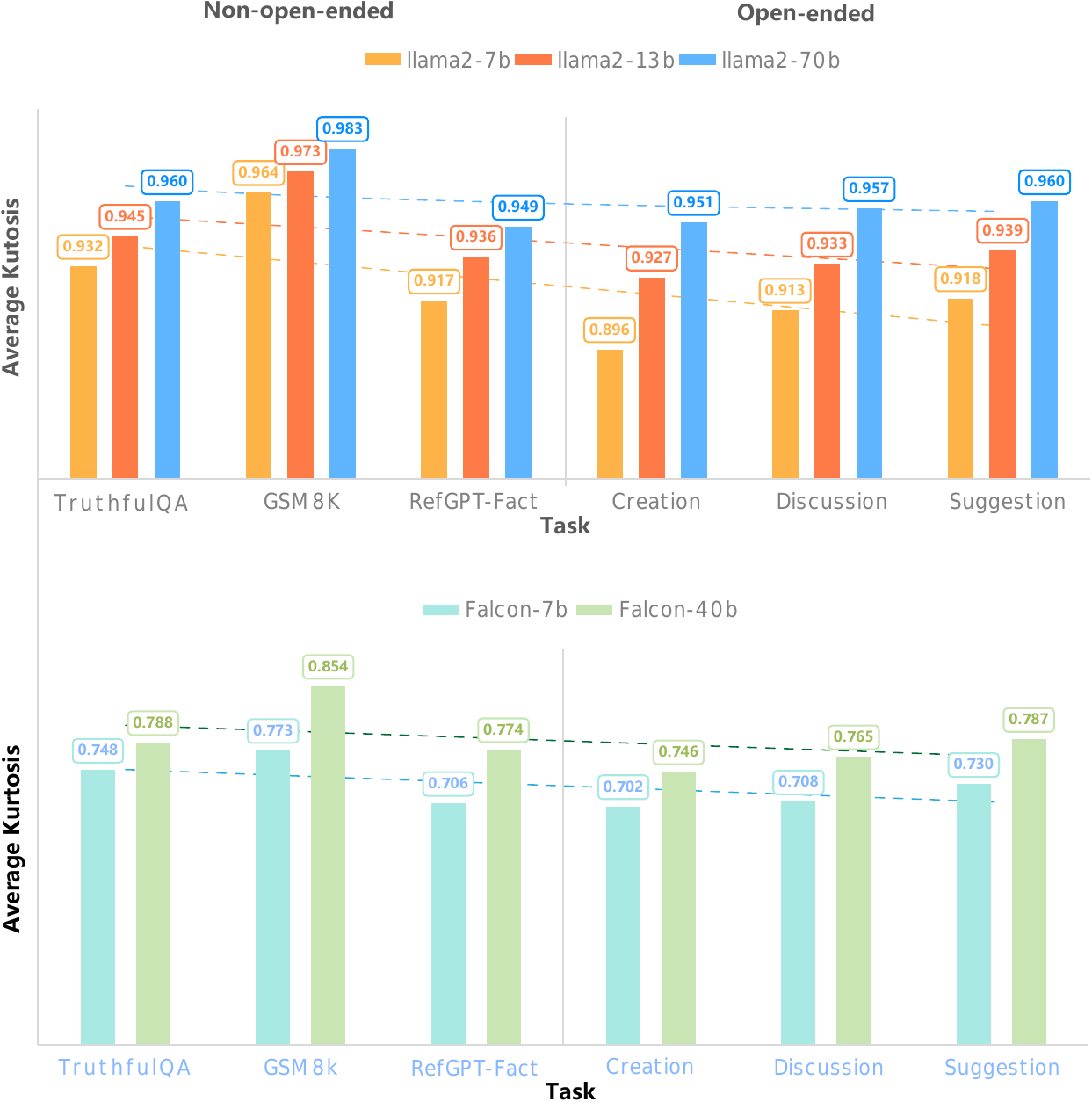}
	\caption{The result of question awareness evaluation. The dotted lines are the trend lines of the kurtosises, which are linearly fitted.}
	\label{fig:QuAT_eval}
\end{figure*}

\subsection{Formulation of the Next Token Prediction}
To better clarify the question awareness, we first give a formulation of the next token prediction in the text generation. For an auto-regressive language model, denoted as $\phi$, given a question $x$, we can calculate the output distribution of the next token $\hat{y}_t$ as follows: 
\begin{equation}
\label{eq:softmax}
\begin{gathered}
     p_{\phi} (\hat{y}_t|x, y_{<t}) = \mathrm{Softmax}\left(\frac{l_{\phi, t}(x, y_{<t})}{\mathcal{T}}\right),
\end{gathered}
\end{equation}
where $l_{\phi, t}(x, y_{<t})$ is the output logit of the token at the step $t$ and $\mathcal{T}$ is the temperature of the Softmax function \citep{bridle1989training}. We sample from the output distribution $p_{\phi} (\hat{y}_t|x, y_{<t})$ to generate the next token. For the temperature $\mathcal{T}$, we can consider the original Softmax function without $\mathcal{T}$ as the Softmax function with a temperature of 1. As shown in Figure \ref{fig:cover}, if we sample the next token with a lower temperature, the output distribution will get steeper thus more likely sampling the token with a large probability. Therefore, the temperature influences the kurtosis of the output distributions and externally changes the question awareness of LLMs. In common practice, we tune the temperature, which is a hyperparameter, to decide how deterministic LLMs should be to answer the question. We usually select a fixed temperature and will not frequently change the value because it is tedious to tune for an optimal temperature for every question. 



\subsection{Metric}
The steepness of the next-token distribution, $p_{\phi} (\hat{y}_t|x, y_{<t}) = (p_1, p_2, \dots, p_n)$, where $n$ stands for vocabulary size, indicates how deterministic the LLMs are, reflecting the question awareness. To measure the steepness, we introduce kurtosis as the metric. If the distribution is steeper, the model is more deterministic on this generated token and the kurtosis gets larger. We use the average kurtosis of the distributions of all answer tokens to reflect the general determinacy of the answer. \textbf{To simplify the illustration, in this paper, we consider the average kurtosis and question awareness to be the same thing.} We calculate the average kurtosis $\mathcal{K}$ as follows:
\begin{equation}
\label{eq:kurt}
\begin{gathered}
\kappa_t = \frac{\frac{1}{n}\sum^{n}_{i=1}(p_i-\overline{p})^4}{\left(\frac{1}{n}\sum^{n}_{i=1}(p_i-\overline{p})^2\right)^2}-3, \\
\mathcal{K} = \frac{1}{T}\sum^T_{t=1}(\kappa_t / \kappa_{\mathrm{one-hot}}),
\end{gathered}
\end{equation}
where $\kappa_t$ is the kurtosis of the distribution of the token at step $t$. For the discrete distribution, the value of kurtosis is related to the value $n$. As the vocabulary sizes of LLMs are different, we have to normalize the kurtosis for fair comparison. Because the one-hot distribution is the steepest and has the largest kurtosis, we divide the kurtosis by $\kappa_{\mathrm{one-hot}}$ to normalize the kurtosis to $(0,1)$.  

\subsection{Evaluation Process}
We first evaluate the essential question awareness in LLMs by calculating the average kurtosises with the default temperature of 1. We then explore the influence of question awareness on the performance by externally adjusting average kurtosises (question awareness) using different temperatures.



\begin{figure*}[ht]
	\centering
	\includegraphics[width=1.0\textwidth]{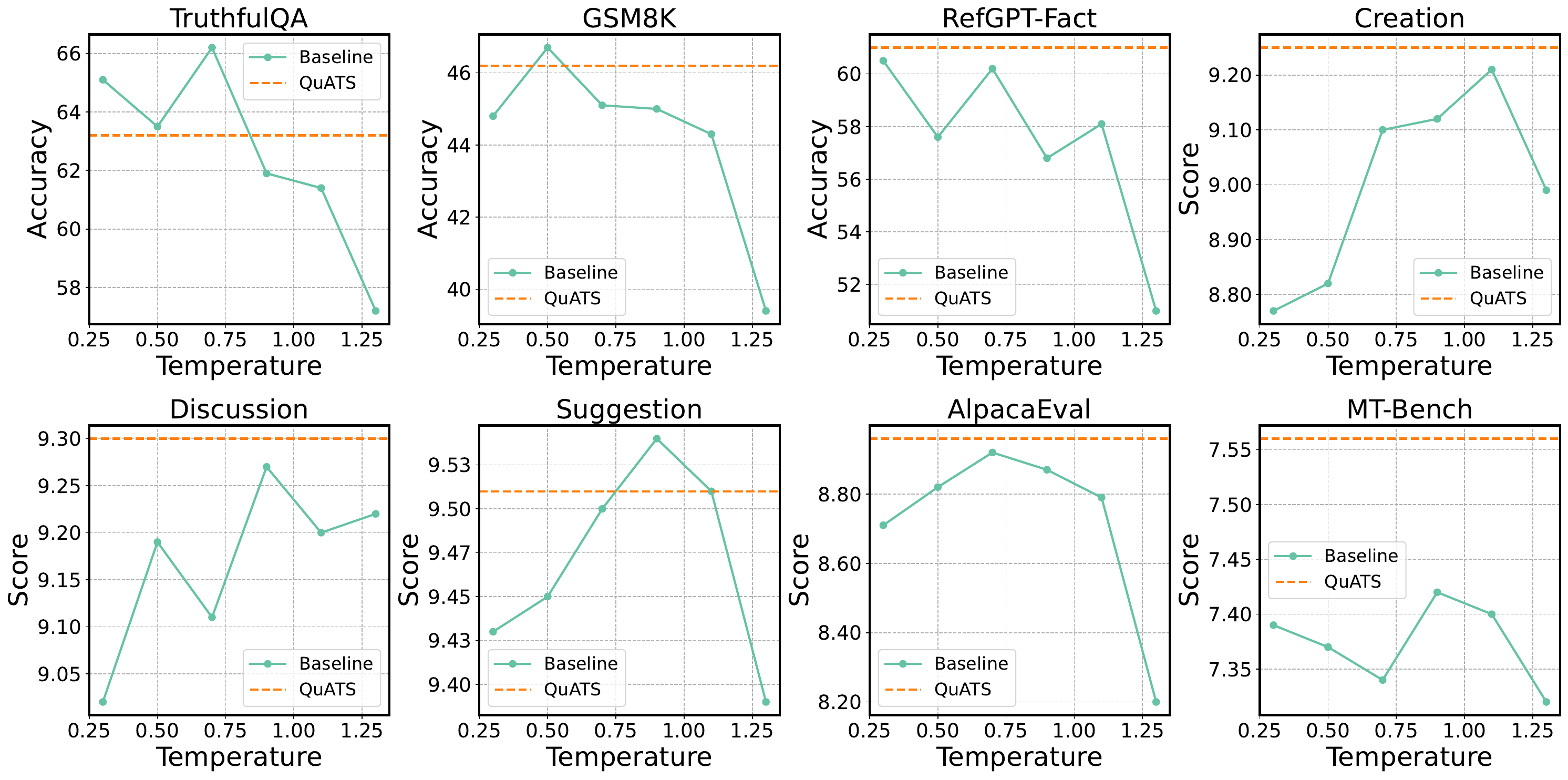}
	\caption{Comparison between QuATS and baselines with different fixed temperatures using LLaMA 2-Chat 13B \citep{touvron2023llama} on downstream tasks. The temperatures adjust the kurtosises, which influence the performance in open-ended and non-open-ended questions differently. In contrast, the adaptive temperature strategy of QuATS consistently outperforms temperature sampling with fixed temperatures.}
	\label{fig:ablation}
\end{figure*}

To evaluate question awareness, we construct an evaluation dataset where questions have distinctions in terms of the determinacy to answer them. Therefore, we collect the questions of mainly two types, non-open-ended and open-ended questions. We collect three types of non-open-ended questions that have only fixed/limited answers: (1) TruthfulQA \citep{lin2022truthfulqa}: questions about commonsense knowledge. (2) GSM8K \citep{cobbe2021training}: school math word problems. (3) RefGPT-Fact \citep{yang2023refgpt}: questions about world knowledge, including factual knowledge of histories, celebrities, places, and so on. We also collect open-ended questions that encourage more creative answers: (1) Creation: content creation including writing articles, emails, and so on. (2) Discussion: discussion on a certain topic. (3) Suggestion: offering useful suggestions. All these subsets of non-open-ended type are carefully selected by humans from ShareGPT dataset \citep{dom2023sharegpt}. We evaluate chat models with different sizes, including LLaMA 2-Chat 7B/13B/70B \citep{touvron2023llama}, Falcon-instruct 7B/40B \citep{penedo2023refinedweb}. It is noted that we do not evaluate closed-source models like GPT-4 \citep{openai2023gpt4} because we can not obtain the output distributions from the APIs.

To investigate the influence of question awareness on the performance, we evaluate the performance of LLaMA 2-Chat 13B \cite{touvron2023llama} on the evaluation dataset using different temperatures for generation. Details about the metric of open-ended questions in Figure \ref{fig:ablation} are introduced in Sec \ref{sec:eval}. 


\subsection{Results and Analysis}
\paragraph{LLMs lack a strong sense of question awareness.} In Figure \ref{fig:QuAT_eval}, according to the trend lines, the kurtosises of non-open-ended questions are not significantly higher than the ones of open-ended questions in most models. For non-open-ended questions, LLMs have fundamental question awareness, e.g., answering commonsense knowledge in TruthfulQA and math problems in GSM8K. However, LLMs do not show more determinacy when answering questions about factual knowledge in RefGPT-Fact, where the kurtosises are close to the average of open-ended questions. It shows that LLMs sometimes struggle to recognize questions about world knowledge that require careful answers, thus easily leading to casual and hallucinated answers. For open-ended questions, similar problems can be found: Most LLMs have relatively lower kurtosis in Creation but fail to be more creative and casual in Discussion and Suggestion. It suggests the models may give repetitive answers to these questions if we ask several times. 

\paragraph{Question awareness greatly affects model performance.}
In Figure \ref{fig:ablation}, for the non-open-ended questions, the results (green lines) show that the model has better performance with relatively low temperatures (steeper distributions) and the performance decreases as the temperatures get higher. It indicates LLM is not determinant enough (with a default temperature of 1) and lacks question awareness essentially. Therefore, if we increase the steepness with a lower temperature, it improves the performance for non-open-ended questions. Opposite conclusions for open-ended questions can be also observed.

\paragraph{Larger models have more confidence in text generation.}
Though we do not observe an emergence of question awareness in larger models, we find that models with larger sizes tend to be more deterministic and focused with higher kurtosis. It means they are more confident in their answers.

\begin{figure*}[ht]
\begin{minipage}{.52\textwidth}
	\centering
	\includegraphics[width=1.0\textwidth]{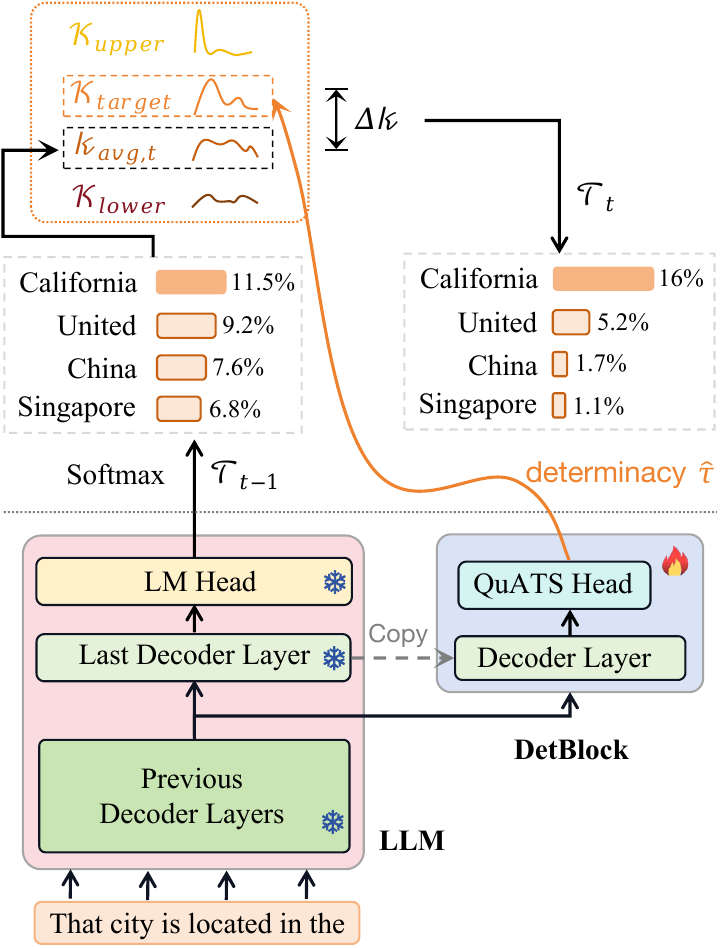}
	\caption{The overview of the QuATS.}
	\label{fig:overview}
\end{minipage}
\hfill
\begin{minipage}{.44\textwidth}
\begin{algorithm}[H]
\caption{QuATS in the inference} 
\label{al:QuATS}
\hspace*{0.02in} {\bf Input:} 
hidden states $h_\phi(x)$, output logits $l_{\phi, t}(x, y_{<t})$, kurtosis mean $\mathcal{K}_{avg}$ and std $\mathcal{K}_{std}$\\
\hspace*{0.02in} {\bf Output:} answer sequence $y$
\begin{algorithmic}[0]
\State $\hat{\tau} = \mathrm{DetBlock}(h_{\phi}(x))$ 
\State $\mathcal{K}_{upper} = \mathcal{K}_{avg} + \lambda \cdot \mathcal{K}_{std}$, 
\State $\mathcal{K}_{lower} = \mathcal{K}_{avg} - \lambda \cdot \mathcal{K}_{std}$ 
\State $\mathcal{K}_{target} = \hat{\tau} \cdot (\mathcal{K}_{upper} - \mathcal{K}_{lower}) + \mathcal{K}_{lower}$ 
\State $t=1$, ${\mathcal{T}_0} = 1.0$, $y = [\;]$ 
\Repeat
    \State $\hat{p_{\phi}} (\hat{y}_t|x, y_{<t}) = \mathrm{Softmax}\left(\frac{l_{\phi, t}(x, y_{<t})}{\mathcal{T}_{t-1}}\right)$ 
    \State $\kappa_t = \frac{\frac{1}{n}\sum^{n}_{i=1}(p_i-\overline{p})^4}{\left(\frac{1}{n}\sum^{n}_{i=1}(p_i-\overline{p})^2\right)^2)}-3$ 
    \State $\kappa_{avg, t} = \frac{1}{t}\sum^t_{i=1}\kappa_i$
    \State $\hat{\mathcal{T}}_{t} =1 + \eta \cdot (\kappa_{avg, t} - \mathcal{K}_{target})t$ 
    \State $ \hat{\mathcal{T}}_{t} = \mathrm{Clamp}(\hat{\mathcal{T}}_{t}, \mathcal{T}_{min}, \mathcal{T}_{max})$
    \State $p_{\phi} (\hat{y}_t|x, y_{<t}) = \mathrm{Softmax}\left(\frac{l_{\phi, t}(x, y_{<t})}{\mathcal{T}_{t}}\right)$ 
    \State $\hat{y_t} = \mathrm{Sample}(p_{\phi} (\hat{y}_t|x, y_{<t}))$ 
    \State $y = \mathrm{Append}(y, \;\hat{y_t})$ 
    \State $t = t + 1$
\Until $\hat{y_t} == \mathrm{<|endoftext|>}$ 
\\
\Return $y$
\end{algorithmic}
\end{algorithm}
\end{minipage}
\end{figure*}

\section{Question Awareness Temperature Sampling}
Based on the findings above, we propose the Question Awareness Temperature Sampling (QuATS) to enhance the question awareness of LLMs by an adaptive temperature strategy, which greatly improves the model performance.



\subsection{Training A DetBlock to Predict Determinacy}
It is a challenge that temperature is a hyperparameter that can not be optimized. We bypass the direct optimization and use the neural network to predict the tendency of how temperature changes according to the determinacy.
We introduce a tiny network called \textbf{DetBlock} to predict the determinacy and leverage it to find the optimal temperature for sampling. Before doing inference with QuATS, we train the DetBlock to predict how deterministic and focused LLMs should be based on the given questions. After DetBlock is ready, we convert the predicted determinacy score to the temperature and adaptively adjust the temperature on the fly during inference.

\paragraph{Training Dataset}
To train the DetBlock, we construct a dataset where questions are rated by determinacy scores based on the artificial criteria. To be specific, we rate open-ended questions requiring less determinacy with lower determinacy scores and vice versa. We use the questions as the input and determinacy scores as training labels.

\paragraph{DetBlock Structure}

As shown in Figure \ref{fig:overview}, we design a tiny network to be DetBlock to predict the determinacy score. The backbone of DetBlock is copied from the last decoder layer of the LLM. We add the QuATS head to the end of the backbone to predict a scalar score of determinacy. 

\paragraph{Training Process} We collect the penultimate hidden states of the question $x$, denoted as the $h_{\phi}(x)$. We feed the $h_{\phi}(x)$ to the DetBlock to predict the determinacy score $\tau$ by minimizing the Mean Square Error (MSE) loss as follows:
\begin{equation}
\label{eq:score}
     \hat{\tau} = \mathrm{DetBlock}(h_{\phi}(x)),
\end{equation}
\begin{equation}
    \mathcal{L}_{QuATS}(\phi) = \frac{1}{2}(\tau - \hat{\tau})^2.
\end{equation}
During the training of DetBlock, we \textbf{freeze} the weights of the LLM so that the performance of the original model will not be affected.

Besides that, we need to record the mean and standard deviation of the kurtosis of the output distributions during training, denoted as $\mathcal{K}_{avg}$ and $\mathcal{K}_{std}$. We record these values for the inference later. We calculate the $\mathcal{K}_{avg}$ and $\mathcal{K}_{std}$ using the exponential moving average as follows:
\begin{equation}
\label{eq:avg_std}
    \begin{gathered}
        \mathcal{K}_{avg, s} = \beta \cdot \mathcal{K}_{avg, s-1} + (1 - \beta) \cdot \hat{\mathcal{K}}_{avg, s},\\
        \mathcal{K}_{std, s} = \beta \cdot \mathcal{K}_{std, s-1} + (1 - \beta) \cdot \hat{\mathcal{K}}_{std, s},
    \end{gathered}
\end{equation}
where the $\hat{\mathcal{K}}_{avg, s}$ and $\hat{\mathcal{K}}_{std, s}$ are calculated by averaging the means and standard deviations of kurtosis of the whole batch at training step $s$. 


\subsection{Inference with QuATS}
Before sampling the next token in the inference, we use DetBlock to predict the determinacy score $\hat{\tau}$ in Eq \ref{eq:score} from the input question. If the determinacy score is large, it means the LLMs are required to be more deterministic to answer this question. The prediction of the determinacy score will be done only once at the start of the generation.

Though we can rescale the determinacy score to get the temperature, it is noted that predicting temperature in this way does not take into account the intrinsic question awareness of LLMs. Based on the question awareness evaluation in Sec \ref{sec:QuAT_eval}, we observe that LLMs have fundamental question awareness in some cases, which means some output distributions are steep/flat enough to give a deterministic/creative answer. If we directly change the temperature, it may lead to overcorrection. Therefore, to avoid overcorrection, we propose QuATS to dynamically adjust the temperature of every decoded token based on both the determinacy score and original output distributions.

To implement QuATS in the inference, we calculate three things step by step: (1) target kurtosis $\mathcal{K}_{target}$, (2) current average kurtosis of the answer $\kappa_{avg}$, and finally (3) estimated temperature $\mathcal{T}$. We predict the temperature for every token to be decoded by projecting $\kappa_{avg}$ to $\mathcal{K}_{target}$.

\paragraph{Target Kurtosis} 
We want to correct the output distribution to be steeper or flatter according to the question. Therefore, we have to find out the target kurtosis we want the distribution to have. The target kurtosis takes the value from the kurtosis interval $[\mathcal{K}_{lower}, \;\mathcal{K}_{upper}]$ as follows:
\begin{equation}
\label{eq:interval}
    \begin{gathered}
        \mathcal{K}_{upper} = \mathcal{K}_{avg} + \lambda \cdot \mathcal{K}_{std}, \\
        \mathcal{K}_{lower} = \mathcal{K}_{avg} - \lambda \cdot \mathcal{K}_{std},
    \end{gathered}
\end{equation}
where the $\mathcal{K}_{avg}$ and $\mathcal{K}_{std}$ are recorded in Eq \ref{eq:avg_std} when training the DetBlock. The kurtosis interval represents the range that the kurtosis of the model output distribution can commonly reach. According to the kurtosis interval, we use the predicted determinacy score $\hat{\tau}$ from DetBlock to calculate a target kurtosis $\mathcal{K}_{target}$ proportionately from the interval as follows:
\begin{equation}
\label{eq:k_target}
    \mathcal{K}_{target} = \hat{\tau} \cdot (\mathcal{K}_{upper} - \mathcal{K}_{lower}) + \mathcal{K}_{lower},
\end{equation}
The target kurtosis $\mathcal{K}_{target}$ lies in the kurtosis interval with $0 \leq \hat{\tau} \leq 1$. It constrains the range of the kurtosis of adjusted output distributions, which avoids overcorrection that the adjusted distributions are too steep or too flat.

\paragraph{Current Average Kurtosis}
Our next goal is to calculate the current average kurtosis of the answer so that we can know the starting point to be projected to target kurtosis. We use the mean of the kurtosises of the decoded token distributions to represent this kurtosis:
\begin{equation}
\label{eq:k_avg}
    \kappa_{avg, t} = \frac{1}{t}\sum^t_{i=1}\kappa_i,
\end{equation}
The $\kappa_{avg, t}$ is a running mean which is updated as the number of decoded tokens increases. We use the running mean to approximate it because we can not know the kurtosis of the whole output distribution before generation ends. Therefore, as the step $t$ increases, the running mean $\kappa_{avg, t}$ will be approximate to the true average kurtosis of the whole answer distribution.

\begin{figure*}[ht]
	\centering
	\includegraphics[width=0.95\textwidth]{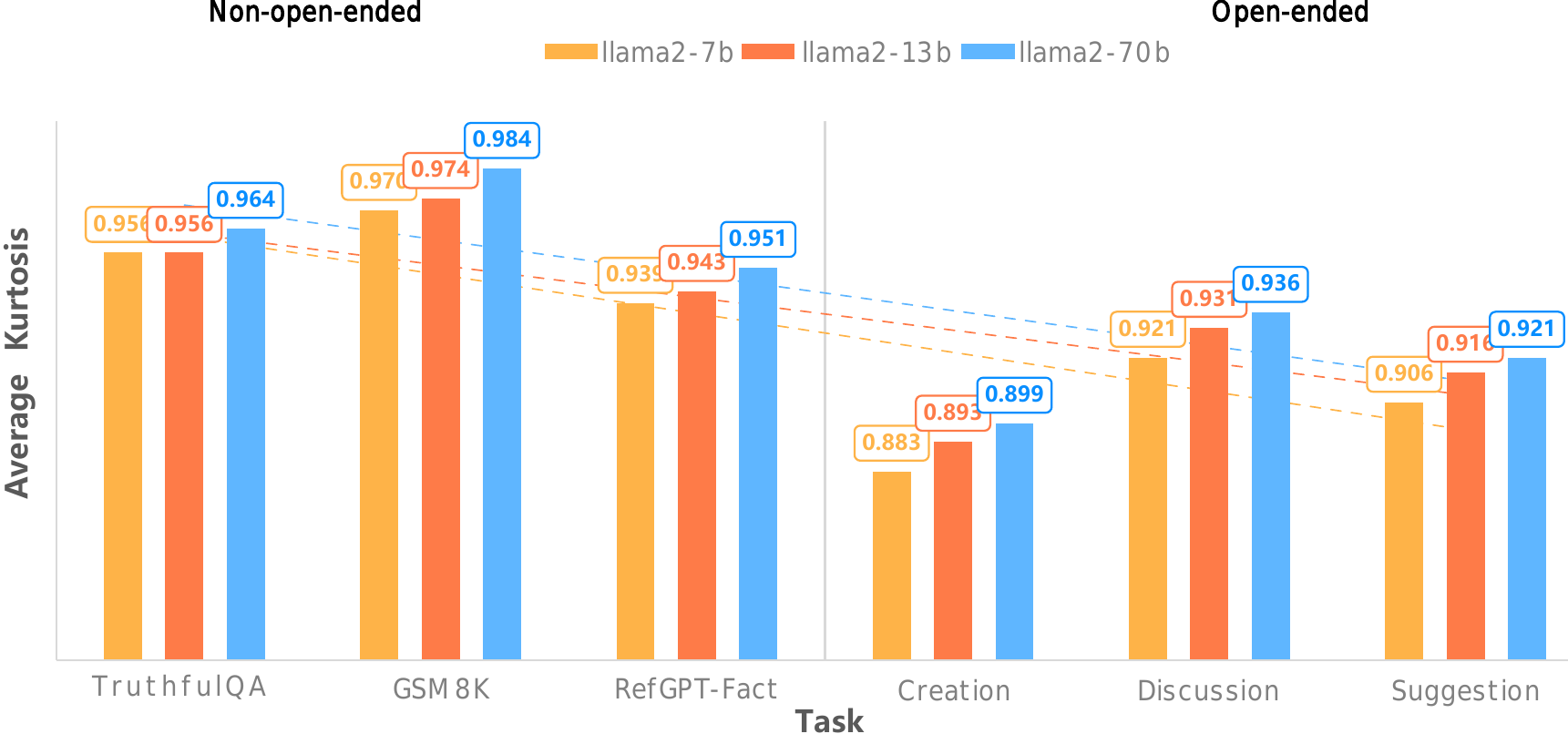}
	\caption{The result of question awareness evaluation of LLaMA 2-Chat models using the QuATS.}
	\label{fig:QuAT_eval_after}
\end{figure*}

\begin{table*}[ht]
\centering
\setlength{\tabcolsep}{3.85pt}
\caption{Evaluating the performance of LLMs using QuATS on various benchmarks. Acc represents the accuracy and Sco represents the score, which is rated according to LLM-as-a-judge in MT-Bench \citep{zheng2023judging}.}
\label{tab:results}
 \small
{
\begin{tabular}{l|ccc|ccc|cc}
\toprule
\multirow{3}{*}{\textbf{Model}} &
\multicolumn{3}{c}{Non-open-ended} & \multicolumn{3}{c}{Open-ended} & \multicolumn{2}{c}{Conversation} \\
  &\textbf{TruthfulQA} & \textbf{GSM8K} & \textbf{RefGPT-Fact} & \textbf{Creation} & \textbf{Discussion} & \textbf{Suggestion} & \textbf{AlpacaEval} & \textbf{MT-Bench}  \\
& Acc & Acc & Acc & Sco & Sco & Sco & Sco  & Sco\\
\midrule
$\mathrm{\textbf{LLaMA 2 7B}}$  
& 50.1 & 21.0 & 51.1 & \textbf{9.19} & 9.35 & 9.40 & 8.57 & 6.88 \\
\quad + QuATS         
& \textbf{55.3} & \textbf{29.8} & \textbf{56.3} & 9.07 & \textbf{9.35} & \textbf{9.43} & \textbf{8.71} & \textbf{7.19} \\
\midrule
$\mathrm{\textbf{LLaMA 2 13B}}$      
& 62.4 & 43.0 & 58.4 & 9.22 & 9.26 & \textbf{9.55} & 8.81 & 7.43 \\
\quad + QuATS               
& \textbf{63.2} & \textbf{46.2} & \textbf{61.0} & \textbf{9.25} & \textbf{9.30} & 9.51 & \textbf{8.96} & \textbf{7.56} \\
\midrule
$\mathrm{\textbf{LLaMA 2 70B}}$  
& 59.2 & \textbf{62.7} & 66.1 & \textbf{9.33} & 9.48 & 9.49 & 9.20 & 7.78 \\
\quad + QuATS             
& \textbf{61.9} & 61.5 & \textbf{68.5} & 9.29 & \textbf{9.51} & \textbf{9.52} & \textbf{9.24} & \textbf{7.83} \\
\midrule
\midrule
$\mathrm{\textbf{Falcon 7B}}$       
& 26.1 & 2.1 & 28.8 & 6.21 & 6.28 & 6.61 & 5.45 & 4.50 \\
\quad + QuATS               
& \textbf{32.7} & \textbf{2.9} & \textbf{33.4} & \textbf{6.41} & \textbf{6.54} & \textbf{6.72} & \textbf{5.82} & \textbf{5.11}\\
\midrule
$\mathrm{\textbf{Falcon 40B}}$       
& 50.2 & 13.6 & 46.2 & 7.33 & 7.91 & \textbf{8.21} & 7.26 & 6.30 \\
\quad + QuATS               
& \textbf{53.0} & \textbf{15.3} & \textbf{50.3} & \textbf{7.57} & \textbf{8.06} & 8.16 & \textbf{7.42} & \textbf{6.59} \\
\bottomrule
\end{tabular}
}
\end{table*}

\paragraph{Estimated Temperature}
By changing the temperature of the Softmax function, we can adjust the distribution to project the average kurtosis $\kappa_{avg, t}$ of the answer to the target kurtosis $\mathcal{K}_{target}$. For the generation step $t$, we estimate the temperature as follows:
\begin{equation}
\label{eq:adjust_temp}
      \hat{\mathcal{T}}_{t} =1 + \eta \cdot (\kappa_{avg, t} - \mathcal{K}_{target})t, 
\end{equation}
\begin{equation}
\label{eq:clamp}
        \hat{\mathcal{T}}_{t} = \mathrm{Clamp}(\hat{\mathcal{T}}_{t}, 
        \mathcal{T}_{min}, \mathcal{T}_{max}).  
\end{equation}
In Eq \ref{eq:adjust_temp}, the temperature in QuATS is decided by three factors: (1) the difference between $\kappa_{avg, t}$ and $\mathcal{K}_{target}$, (2) the generation step $t$, (3) a coefficient $\eta$ to control the adjustment speed. For the first factor, if $\kappa_{avg, t} > \mathcal{K}_{target}$, it means the current average kurtosis is higher than the target kurtosis, thus we need to increase the temperature to flatten them, and vice versa. For the second factor, as the generation step $t$ increases, the $\kappa_{avg, t}$ tends to approach the true average kurtosis of the whole answer. Thus the $(\kappa_{avg, t} - \mathcal{K}_{target})$ should exert a greater impact on the temperature adjustment. We need to clamp the temperature between an interval to avoid being too high or too low in Eq \ref{eq:clamp}.



\section{Experiment}
In this section, we conduct experiments to showcase that QuATS can enhance question awareness using the adaptive temperature strategy and consistently improve the model performance.

\subsection{Evaluation Setup}
\label{sec:eval}
To verify the effectiveness of the QuATS, we evaluate if LLMs with QuATS have a better awareness of different question types and better performance on our question awareness evaluation dataset in Sec \ref{sec:QuAT_eval}. Besides that, we choose two LLM benchmarks, namely AlpacaEval \citep{alpaca_eval} and MT-Bench \citep{zheng2023judging}, which test if the models with QuATS can handle conversations of different scenarios. We set models with a temperature of 1 as the baselines. To check the answers to open-ended questions, we follow the official implementation of LLM-as-a-judge from the MT-Bench\footnote{https://github.com/lm-sys/FastChat/tree/main/fastchat} and use GPT-4 turbo as a judge to score 1 to 10 for the answers, as shown in Table \ref{tab:results}.

\subsection{Results and Analysis}
From Figure \ref{fig:QuAT_eval_after}, we evaluate the question awareness of LLaMA 2-Chat models using QuATS. Compared to the ones in Figure \ref{fig:QuAT_eval}, the descending trend lines have shown a distinction in the awareness between non-open-ended and open-ended questions. The models with QuATS choose to be more deterministic with higher kurtosises in answering non-open-ended questions. Similar findings can be observed in open-ended questions.

For model performance, in Table \ref{tab:results}, we can see that QuATS largely improves the LLM performance in the various tasks, especially in the non-open-ended questions. It means that a better awareness of non-open-ended questions can alleviate the hallucination. For the results of two comprehensive LLM benchmarks, MT-Bench and AlpacaEval, both LLaMA 2 and Falcon have significant improvements over the baselines, which show the QuATS is useful for different models with different sizes on open-domain conversations. We observe that smaller models like LLaMA 2 7B and Falcon 7B have more performance gains than larger models. It can be inferred that the distribution of larger models originally has more appropriate tokens with high probabilities thus the effectiveness of additional adjustment on the steepness of the distribution tends to be smaller.

In Figure \ref{fig:ablation}, we also compare the performance of the model using QuATS with baselines using different fixed temperatures. QuATS consistently outperforms the naive temperature sampling with different fixed temperatures on these tasks.



\section{Related Work}
Controlling text generation in LLMs has seen significant advancements in recent years. Sampling methods play a crucial role in controlling the output quality and diversity of generated text. We introduce temperature sampling and corresponding advanced techniques in text generation.



\paragraph{Temperature Sampling} 
Greedy sampling selects the token with the highest predicted probability, resulting in deterministic and often repetitive text. Random sampling selects tokens based on the probabilities, introducing randomness to alleviate the repetition. We can further adjust the temperature in the Softmax function \citep{bridle1989training} to control the token probabilities. Temperature sampling can be seen as the trade-off between creativity and determinacy in the generated text. Our QuATS adaptively controls the steepness of output distributions by adjusting the temperature.

\paragraph{Post-processing Techniques} 
Because the tokens with higher probabilities are probably appropriate choices, we can choose only to select these tokens, avoiding sampling nonsensical tokens. Top-k sampling \citep{fan2018hierarchical} narrows down the token selection to the top-k most probable tokens, increasing the likelihood of coherent text and balancing diversity and quality. Similar to the motivation of top-k sampling, nucleus sampling \citep{holtzman2020curious}, also known as top-p sampling, dynamically selects the top-p fraction of tokens with the highest probabilities. Locally typical sampling \citep{meister2023locally} posits the abstraction of natural language generation as a discrete stochastic process and samples tokens according to conditional entropy. Entmax sampling \citep{martins2020sparse} leverages entmax transformation to train and sample from a natively sparse language model. Keyword-based sampling \citep{s2023keyword} uses knowledge distillation techniques to extract keywords and samples using these extracted keywords. It is noted that these post-processing techniques are compatible with QuATS because QuATS only adjusts the output distribution, which can be further post-processed.

\section{Conclusion}
In this paper, we highlight the question awareness of LLMs, which receives little attention from previous studies. While LLMs exhibit a fundamental awareness of open-ended and non-open-ended questions, they do falter in certain domains, often leading to casual or inaccurate responses. To bridge the gap, we introduce Question Awareness Temperature Sampling (QuATS), enabling LLMs to autonomously adapt their response determinacy based on question type. Our experiments showcased the efficacy of QuATS, significantly enhancing LLM performance across various benchmarks. 

\section*{Limitations}
In this paper, we explore the question awareness of LLMs from the perspective of output distributions and enhance this ability by adjusting the temperature. However, the question awareness should be the intrinsic ability that the model should have. However, QuATS only improves this ability externally by the DetBlock but does not enhance the model itself. 

We believe the question awareness of LLMs is a valuable subject, providing a new perspective of hallucinations in LLMs. How to improve the intrinsic question awareness to reduce the hallucinations is worthy of exploration for future work.

\bibliography{custom}
\bibliographystyle{acl_natbib}

\appendix
\clearpage
\appendix
\onecolumn
\section{DetBlock Training Details}
To train a DetBlock, we collect a training dataset consisting of 3.5K high-quality questions, which are carefully selected by humans from the ShareGPT dataset \citep{dom2023sharegpt}. Using the criteria shown in Table \ref{tab:train_prompt}, we manually label the questions with the determinacy scores according to how deterministic the answers should be. The questions are rated by two persons separately from not very deterministic (1 point) to highly deterministic (4 points). We average the two scores and rescale the averaged score to (0, 1) as the final determinacy score.

We train the DetBlock based on LLaMA 2-Chat 7B/13B/70B \citep{touvron2023llama} and Falcon-instruct 7B/40B \citep{penedo2023refinedweb}. We train for 2 epochs with a batch size of 32 on the 7B models with a learning rate of 2e-5, the 13B model with 1e-5, and the 40B/70B models with 5e-6.



\begin{table*}[h]
\centering
\def\arraystretch{1.5}
\caption{The criteria for rating the determinacy score in the training dataset.}
\label{tab:train_prompt}
\begin{tabular}{m{4cm}|m{10cm}}
\toprule
\makecell{Highly Deterministic\\ (4 points)} & Questions/instructions that have a unique answer, including mathematical calculations and factual knowledge. \\
\midrule
\makecell{Fairly Deterministic\\ (3 points)} & Questions/instructions related to logical reasoning, code modification and creation, text rewriting and summarization, text translation, reading comprehension. \\
\midrule
\makecell{Moderately Deterministic\\ (2 points)} & Questions/instructions related to code discussions and creative inquiries that require a certain level of expertise. \\
\midrule
\makecell{Not Very Deterministic\\ (1 point)} & Creative and open-ended questions/instructions (e.g., "What do you think about...?" "How do you see...?").\\
\bottomrule
\end{tabular}
\end{table*}

\end{document}